\documentclass[10pt,twocolumn,letterpaper]{article}

\usepackage[pagenumbers]{cvpr} 

\usepackage{bbding}  
\usepackage{pifont}  

\definecolor{gred}{rgb}{0.859,0.267,0.216}
\definecolor{ggreen}{rgb}{0.059,0.616,0.345}
\definecolor{deepblue}{HTML}{27a2c3}

\definecolor{deepred}{HTML}{fe7b5b}
\definecolor{red}{HTML}{FF5B3B}

\definecolor{cvprblue}{rgb}{0.21,0.49,0.74}
\usepackage[pagebackref,breaklinks,colorlinks,allcolors=cvprblue]{hyperref}
\usepackage{multirow}
\usepackage{amsmath}
\usepackage{algorithm}
\usepackage{algorithmic}
\usepackage[accsupp]{axessibility}  
\def\paperID{8666} 
\def\confName{CVPR}
\def\confYear{2025}

\title{AffordDP: Generalizable Diffusion Policy with Transferable Affordance\\ }

\author{Shijie Wu$^{\dagger}$\quad  Yihang Zhu$^{\dagger}$ \quad Yunao Huang\quad Kaizhen Zhu\quad Jiayuan Gu\quad Jingyi Yu\quad \\Ye Shi\thanks{Corresponding authors. $\dagger$ Equal contributions.}\quad  Jingya Wang\footnotemark[1] \\
ShanghaiTech University, Shanghai, China\\
{\tt\footnotesize \{wushj12023,zhuyh2023,huangya2024,zhukzh2024,gujy1,yujingyi,shiye,wangjingya\}@shanghaitech.edu.cn} \\
{\small {\url{https://afforddp.github.io/}}}
}

\begin{document}
\newcommand{\ourwebsite}[0]{\href{https://affordd\Phip.github.io/
}{AffordDP.github.io}}

\twocolumn[{%
\renewcommand\twocolumn[1][]{#1}%
\maketitle
\vspace{-2em}
\begin{center}
    \centering
    \captionsetup{type=figure}
    \vspace{-1em}
    \includegraphics[width=\linewidth]{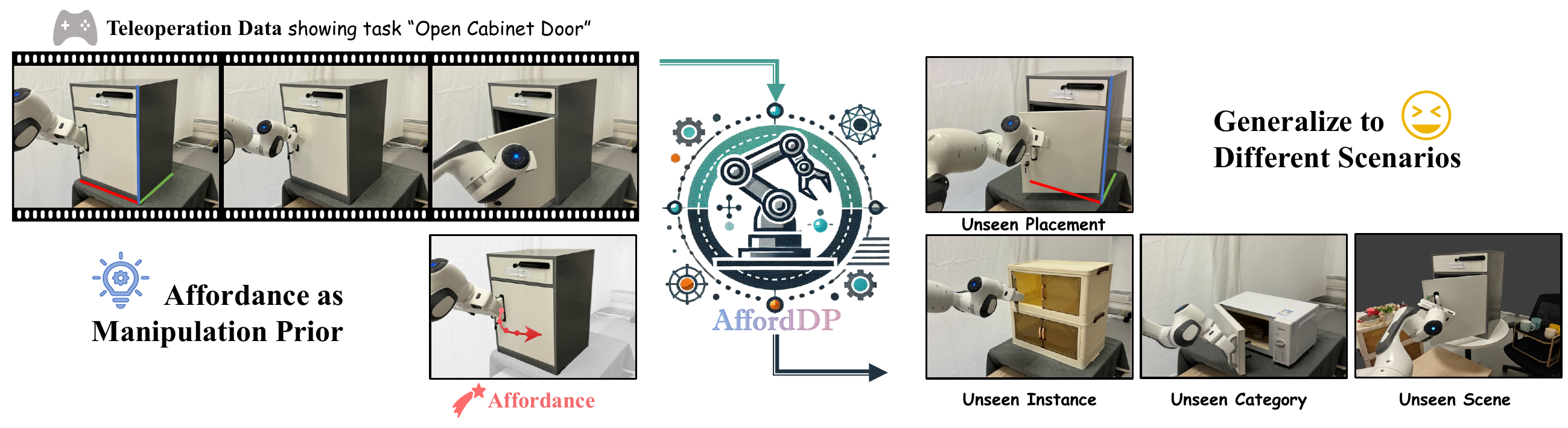}
    \vspace{-1em}
    \captionof{figure}{We propose AffordDP, a novel diffusion-based imitation learning method for generalizable robotic manipulation. AffordDP leverages rich manipulation priors from transferable affordances to enhance generalization to unseen scenarios and incorporates affordance guidance to enable precise control.}
    \label{fig:teaser}
\end{center}
}]
\def\thefootnote{$\dagger$}\footnotetext{Equal contributions.}
\def\thefootnote{*}\footnotetext{Corresponding authors.}

\begin{abstract}

Diffusion-based policies have shown impressive performance in robotic manipulation tasks while struggling with out-of-domain distributions. Recent efforts attempted to enhance generalization by improving the visual feature encoding for diffusion policy. However, their generalization is typically limited to the same category with similar appearances.
Our key insight is that leveraging affordances—manipulation priors that define ``where" and ``how" an agent interacts with an object—can substantially enhance generalization to entirely unseen object instances and categories. We introduce the Diffusion Policy with transferable Affordance (AffordDP), designed for generalizable manipulation across novel categories. AffordDP models affordances through 3D contact points and post-contact trajectories, capturing the essential static and dynamic information for complex tasks. The transferable affordance from in-domain data to unseen objects is achieved by estimating a 6D transformation matrix using foundational vision models and point cloud registration techniques. 
More importantly, we incorporate affordance guidance during diffusion sampling that can refine action sequence generation. This guidance directs the generated action to gradually move towards the desired manipulation for unseen objects while keeping the generated action within the manifold of action space. Experimental results from both simulated and real-world environments demonstrate that AffordDP consistently outperforms previous diffusion-based methods, successfully generalizing to unseen instances and categories where others fail. 
\end{abstract}     

\section{Introduction}
Imitation learning offers an efficient solution for teaching robots a variety of skills by replicating behavior from experts without the need to explicitly program actions or model the environment. It has garnered significant interest across a wide range of fields, including autonomous driving~\cite{hu2023planning, gu2023vip3d, li2022bevformer}, robotic manipulation~\cite{Chi-RSS-23, haldar2023teach, wang2023mimicplay, brohan2022rt, lv2025spatial, wu2025momanipvla, jia2024lift3d}, and legged locomotion~\cite{peng2020learning, pan2020zero, li2023learning}. Recently, diffusion-based imitation models~\cite{Chi-RSS-23, reuss2023goal, ha2023scaling} have gained significant attention within the field. These models conceptualize a robot’s visuomotor policy as a conditional diffusion denoising process, facilitating the learning of conditional action distributions.

Despite their high expressiveness and flexibility, diffusion-based imitation models~\cite{Chi-RSS-23, reuss2023goal, ha2023scaling} still struggle to generalize to out-of-domain distributions,  which significantly limits their practical deployment.
Recently, some pioneering works have sought to enhance the generalization ability of visual feature encoding and directly apply these advancements to diffusion policy \cite{Chi-RSS-23}. For instance, Ze et al.~\cite{Ze2024DP3} utilized 3D point cloud representations instead of 2D representations to improve feature encoding. Concurrently, Yang et al.~\cite{yang2024equibot} proposed integrating SIM(3)-equivariant networks to enhance generalization ability. Additionally, Wang et al.~\cite{wanggendp} introduced 3D semantic fields with multi-view input into diffusion policy to further improve its generalization ability. 
While these approaches may enable diffusion models to generalize to different initial locations, appearances, or poses, their generalization is typically limited to the same category with similar appearances. Subsequently, achieving the generalization of diffusion models to unseen instances and unseen categories remains challenging.

As human beings, we, in contrast, can easily transfer skills to objects with different appearances or even different types of objects. For example, once we learn the skill of opening a cabinet, it is easy to apply this skill to opening a microwave or a refrigerator (just as learning to ride a bicycle makes it easy to ride a tricycle). This ability is due to the vast amount of task-specific interaction prior knowledge that humans possess. This interaction prior knowledge at the object level specifically manifests as knowing “where” and “how” an agent should interact with an object to complete a given task, which is often defined as affordances in many studies~\cite{ardon2020affordances, yu2024seqafford}. However, current imitation learning methods typically define affordances as 2D contact points with trajectory, task-specific keyframes, or value maps~\cite{bahl2023affordances, rana2024affordance, zha2021contrastively}. This significantly limits the representation of affordances in terms of “how” an agent should interact with the object. To tackle this key issue, we design a general representation for both static and dynamic affordances and further propose a diffusion policy with affordance transfer, allowing for the transfer of interaction prior knowledge across different instances and categories.

Specifically, we present a new Diffusion Policy with transferable Affordance that we call AffordDP to address complex manipulations of unseen objects and categories. The transferable affordance encompasses both static and dynamic affordances, capturing 3D contact points and post-contact trajectories, thus providing the “where” and “how” information crucial for manipulation tasks. The transferable affordance is derived from in-domain data and applied to unseen objects by estimating a 6D transformation matrix, utilizing foundational vision models and point cloud registration techniques. AffordDP employs the transferable affordance as an informative condition for the diffusion model, thereby enhancing its capability to generalize across varied manipulative scenarios. However, simply integrating transferable affordance as the diffusion condition may lack sufficient accuracy to achieve zero-shot adaptation for unseen robotic manipulation. To address this challenge, we introduce an affordance-guided sampling process for diffusion policy. In particular, the affordance guidance is defined as the distance between the robot's end-effector position and the static affordance. This guidance during the diffusion sampling process enables the generated action sequence to gradually move toward the desired manipulation for unseen objects while keeping the generated action within the manifold of action space. We adaptively apply the affordance guidance for fine-grained control. The novelty and contributions of this paper are outlined as follows. 

\begin{itemize}

\item We introduce AffordDP that leverages transferable affordance for enhanced generalization to unseen objects and categories that cannot be completed by existing diffusion-based methods. 
\item Our framework enables the transfer of both static and dynamic affordances by utilizing foundational vision models, making it possible to represent complex manipulations with limited in-domain data. 
\item We integrate adaptive affordance guidance to our AffordDP that can direct the generated action sequence to gradually move towards the desired manipulation for unseen objects within the manifold of action space. 
\end{itemize}

\begin{figure*}[t]
    \centering
    \includegraphics[width=\textwidth]{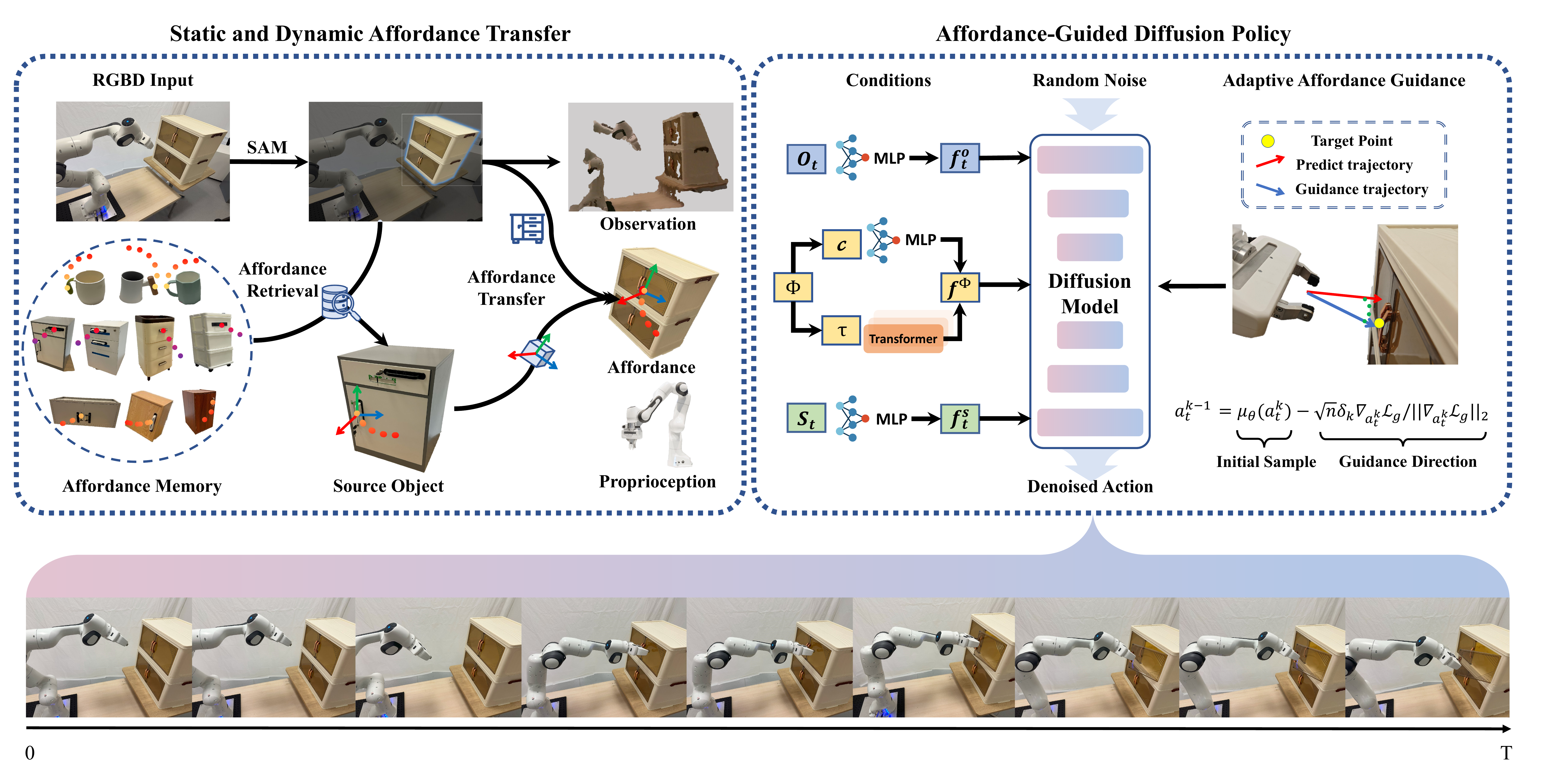}
    \caption{Overview of AffordDP. The left part demonstrates static and dynamic affordance transfer. Given the target scene RGB-D image, AffordDP retrieves a similar object in the affordance memory and transfers its static and dynamic affordance to the target object. The right part illustrates the key components of affordance-guided diffusion policy. Conditioned on 3D affordance, point cloud observation, and robot proprioception, AffordDP utilizes the Diffusion Policy and adaptive affordance guidance for precise control.}
    \label{fig: pipeline}
\end{figure*}

\section{Related work}

\subsection{Diffusion-based Imitation Learning}
As one of the first works~\cite{Chi-RSS-23, reuss2023goal, ha2023scaling} that introduce diffusion into imitation learning, the Diffusion Policy~\cite{Chi-RSS-23} leverages a diffusion denoising process to model conditional action distributions.
Despite its advantages in modeling multimodal data and providing stable training, the Diffusion Policy still faces limitations in generalizing to out-of-domain distributions. When the position or shape of objects changes, the Diffusion Policy struggles to adapt successfully. 3D Diffusion Policy~\cite{Ze2024DP3} addresses this to some extent by using point clouds for visual representation, which improves spatial generalization. Equibot~\cite{yang2024equibot} further advances this by combining SIM(3)-equivariant neural network architectures with diffusion policies, enhancing generalization to SIM(3)-equivariant scenes. Another approach, Im2Flow2Act~\cite{xu2024flow}, combines self-supervised 2D flow prediction with a flow-based Diffusion Policy. However, its reliance on 2D flow limits its effectiveness, failing to capture object movements in the 3D space accurately and leading to less precise actions. Concurrent with our work, GenDP~\cite{wanggendp} and G3Flow~\cite{chen2024g3flow} employ 3D semantic fields to enhance point cloud observations used in Diffusion Policy. However, these approaches require a multi-view setup to generate semantic fields and are limited to category-level generalization. In contrast, our method uniquely conditions diffusion on 3D affordance trajectories, allowing our model to generalize more effectively to objects with similar affordance trajectories, regardless of their category.

\subsection{Generalizable Manipulation for Robotics}
Generalizable manipulation is a principal focus in the field of imitation learning. 
To encourage policies to concentrate on task-relevant factors while minimizing irrelevant distractions, previous methods typically extract object-centric features from images, like object poses~\cite{deng2020self}, task-specific region proposals~\cite{devin2018deep}, keypoints~\cite{manuelli2019kpam} and segmented point clouds~\cite{zhu2023learning}.
A further trend in this field is the integration of affordance into policy learning. 
Mainstream works~\cite{bahl2023affordances, geng2023rlafford, mo2021where2act} tend to learn affordance from large datasets or through intensive reinforcement learning, which can entail significant overhead.
To address these limitations, recent advancements have leveraged semantic information from foundational vision models to enable zero-shot affordance transfer. Pioneering methods such as Robo-ABC~\cite{ju2024robo} and RAM~\cite{kuang2024ram} allow affordance transfer in a training-free manner. However, they are limited to predicting affordances either as 2D contact points or 3D contact points with direction. Additionally, they rely on grasp generators~\cite{fang2023anygrasp} and motion planners~\cite{coleman2014reducing, sundaralingam2023curobo} to perform manipulation tasks, which face challenges in high-dimensional action spaces and complex manipulation tasks. This reliance often results in unimodal actions, limiting effectiveness in environments with significant uncertainty.
Our method advances these developments by combining 2D semantic correspondences with 3D geometrical correspondences, significantly enhancing the capability for effective trajectory-level transfer.

\begin{figure*}[ht]
    \centering
    \includegraphics[width=\textwidth]{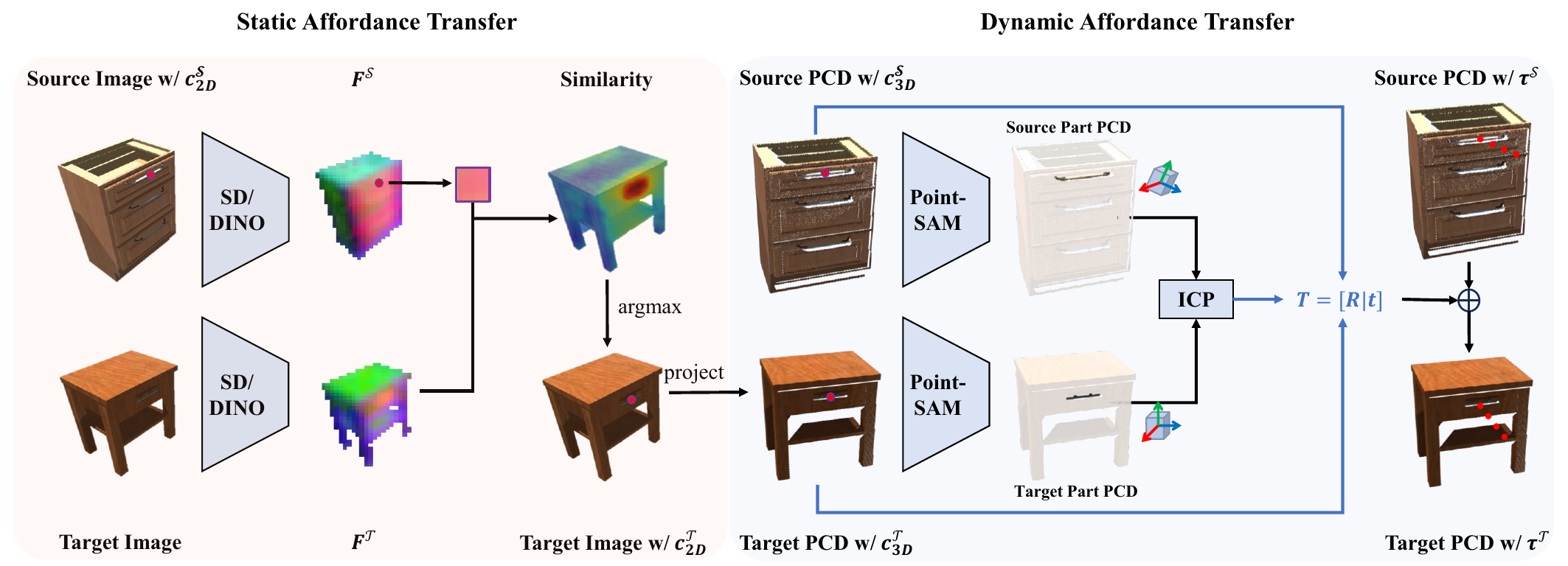}
    \caption{Overview of Static and Dynamic Affordance Transfer. The left part is static affordance transfer, and the right part is dynamic affordance transfer. The source and target images are processed through SD-DINOv2~\cite{zhang2024tale} to generate feature maps $F^S$ and $F^T$. The similarity  is computed and used to find the corresponding target image points using an argmax operation. These points are projected back to obtain point clouds. Point-SAM~\cite{zhou2024point} is then used to obtain the source and target part point clouds. The Iterative Closest Point (ICP) algorithm is applied to the point clouds to determine the transformation matrix $T = [\mathbf{R}|\mathbf{t}]$ for dynamic affordance transfer.}
    \label{fig: afford_transfer}
\end{figure*}

\section{Transfer of Static and Dynamic Affordances} \label{subsec: afford transfer}
AffordDP generates 3D affordances denoted by $\Phi = (c, \tau)$, where $c \in {\mathbb{R}}^3 $ represents the static affordance as 3D contact point and $\tau \in {\mathbb{R}}^{3 \times N}$ represents the dynamic affordance as post-contact trajectory in the form of an ordered list of 3D points. Affordance transfer is defined as generating the affordance for the target object given a source object with its affordance. Upon specifying a target object, AffordDP retrieves the most similar object from the object memory.
The static affordance $c$ is transferred using the semantic correspondence, and the dynamic affordance $\tau$ is transferred using a 6D transformation matrix, whose translation and rotation components are derived independently. An overview is shown in ~\cref{fig: afford_transfer}.\\

\subsection{Affordance memory and similarity retrieval}
To build the affordance memory, we utilize Grounded-SAM~\citep{ren2024grounded} to crop the object from the image and employ the CLIP~\citep{radford2021learning} image encoder to extract the feature vector \( z \). Additionally, we record the contact point as the static affordance \( c \), and the trajectory of the robot end-effector after contacting the object (in world coordinates) as the dynamic affordance \( \tau \). We define the affordance memory \( \mathcal{M} \), which includes the task name \( T \), the affordance \( \Phi = \{c, \tau\} \), the appearance feature \( z \), and the point cloud of the manipulable object \( \mathcal{P} \) recovered from RGB-D images as follows, 
\begin{equation}
    \mathcal{M} = \{(T, \Phi, z, \mathcal{P})\}.
\end{equation}
The cosine similarity between the source and target cropped image feature vectors serves as a proximity metric for retrieval. In practical terms, for an object with task \( T \), we perform retrieval within the subset of the memory corresponding to the same task \( T \).

\subsection{Static affordance transfer}
     Static affordance transfer involves finding the contact point of the target object in 3D world coordinates, given the contact point of the source object. This task is challenging because the two objects may differ in appearance, shape, rotation, or even category. We leverage the rich semantic information provided by foundation vision models to mitigate these variances. We upsample SD-DINOv2~\cite{zhang2024tale} features to generate pixel-level image features. By matching the features of the source static affordance in the 2D image \( c_{2D}^{\mathcal{S}} \) with all features of the target image pixels, we can identify the target static affordance in the 2D image \( c_{2D}^{\mathcal{T}} \). We then back-project the target static affordance \( c_{2D}^{\mathcal{T}} \) to 3D world coordinates as \( c_{3D}^{\mathcal{T}} \).

\subsection{Dynamic affordance transfer}
To transfer dynamic affordances, we utilize a 6D transformation matrix. Previous alignment-based methods that rely on instance-level registration between source and target objects ~\citep{zhu2024vision} often yield unreliable results due to shape and scale discrepancies. For example, transferring a top drawer affordance from a tall closet to a central drawer in a square closet can lead to inaccuracies. To address this issue, we independently estimate the rotation matrix and translation vector, better accommodating structural differences and enhancing the accuracy of affordance transfer.

The translation vector $\mathbf{t}$ is obtained by computing the difference between the source and target 3D contact points. To determine the rotation matrix $\mathbf{R}$, we perform point cloud registration on source and target point clouds. We employ a state-of-the-art promptable 3D segmentation model~\citep{zhou2024point}, using \( c_{3D} \) as a prompt to segment manipulable parts. This approach reduces sensitivity to structural variance, as part point clouds exhibit less variability than complete objects. We then use the iterative closest point (ICP) algorithm~\citep{chen1992object} to register the 3D parts by minimizing the objective function:

\begin{equation}
E(\mathbf{T}) = \sum_{(\mathbf{p}, \mathbf{q}) \in \mathcal{K}} \left[ \left( \mathbf{p} - \mathbf{T} \mathbf{q} \right) \cdot \mathbf{n}_{\mathbf{p}} \right]^2\,,
\end{equation}
where \( \mathbf{T} \) is the transformation, \( \mathcal{K} = \{(\mathbf{p}, \mathbf{q})\} \) is the correspondence set between the target point cloud \( \mathbf{P} \) and the source point cloud \( \mathbf{Q} \), and $\mathbf{n}_{\mathbf{p}}$ is the normal of point $\mathbf{p}$\footnote{We directly apply the implementation of Open3D~\cite{Zhou2018}, correspondence set $\mathcal{K}$ and normal $\mathbf{n}_{\mathbf{p}}$ are also estimated by Open3D.}.

After deriving the rotation matrix \( \mathbf{R} \) from the registration results, we combine it with the translation vector \( \mathbf{t} \) obtained earlier to compute the 6D transformation matrix:

\begin{equation}
\tau^{\mathcal{T}} = \mathbf{R} \tau^{\mathcal{S}} + \mathbf{t}.
\end{equation}
This 6D transformation matrix maps the dynamic affordance of the source object to the target object.

\section{Affordance Guided Diffusion Policy} \label{subsec: afford diffusion}

To enhance the generalization capacity, we design a conditional diffusion model that takes in the transferable affordance as an additional condition, which treats this manipulation prior as a bridge to narrow the gap between training and practical deployment.
However, simply adding affordance as a diffusion condition does not fully leverage its potential, as the diffusion model is inherently focused on modeling the overall action distribution, prioritizing alignment with the distribution rather than adherence to specific task constraints. To address this, we introduce an adaptive affordance-guided sampling process to direct the policy toward static affordance (contact point), enhancing task focus and precision.

\subsection{Conditions for AffordDP}
The overall conditions of our diffusion model are:
\begin{equation}
    \mathcal{C} = \{ O_t, S_t, \Phi\ \},
\end{equation}
where $O_t$ represents the scene point cloud, $S_t$ denotes robot proprioception, and $\Phi$ represents the affordance, which is composed of static affordance $c$ (contact point) and dynamic affordance $\tau$ (trajectory). Our diffusion model builds upon Denoising Diffusion Implicit Models (DDIMs)~\cite{song2020denoising} and aims to approximate affordance-guided conditional distribution $p(a_t|O_t, S_t, \Phi)$, to capture the action distribution through a gradual denoising process from a Gaussian distribution. 

To extract the overall conditioning feature $f_t$ from each element in $\mathcal{C}$, we use distinct encoders. Since only a small region surrounding the robot and target object offers meaningful contextual information, we crop and downsample the scene point cloud before applying an MLP for scene feature extraction, following a similar approach to the 3D Diffusion Policy~\cite{Ze2024DP3}. To capture the full dynamics of the trajectory sequence $\tau$, we utilize a transformer encoder~\cite{NIPS2017_3f5ee243} with a [CLS] token to represent the trajectory feature. The contact point $c$ and robot proprioception $S_t$ features are extracted using separate MLPs. We define the conditioning feature $f_t$ as follows,
\begin{equation}
    f_t = \{f_t^O, f_t^S, f^{\Phi} \},
\end{equation}
where \( f_t^O \) denotes the scene feature, \( f_t^S \) represents the robot state feature, and \( f^{\Phi} \) integrates both the static and dynamic affordance features for simplicity. 

\subsection{Conditional diffusion model training}
Starting from an initial noise vector $a_t^k$, AffordDP uses a noise prediction network ${\epsilon}_\theta$ to iteratively denoise the noise vector conditioned on the conditioning feature $f_t$ till a noise-free action $a_t^0$ is formed. This process can be described using the following equations: 
\begin{equation}
\begin{aligned}
    {a}_t^{k-1} = &\sqrt{\alpha_{k-1}}(a_t^{k} -\sqrt{1-\alpha_k}\epsilon_\theta^k)/ \sqrt{\alpha_k} + \\
    & \sqrt{1-\alpha_{k-1}-\sigma_k^2}\epsilon_\theta^k + \sigma_k \epsilon^k,
\end{aligned}
\end{equation}
\begin{equation}
    \sigma_k = \sqrt{(1-\alpha_{k-1})/(1-\alpha_k)}\sqrt{1-\alpha_k/\alpha_{k-1}}\,,
\end{equation}
where $\epsilon^k$ is random noise, $\alpha_k$ is the noise schedule and $k$ is the number of denoising iterations. Similar to previous diffusion-based policies~\cite{Chi-RSS-23, Ze2024DP3}, we train the noise prediction network ${\epsilon}_\theta$  to predict the noise $\epsilon^k$ added at $k$-th iteration,
\begin{equation}
    \mathcal{L} = ||\epsilon_k-\epsilon_\theta(a_t^k,k, f_t)||^2.
\end{equation}

\subsection{Diffusion sampling with affordance guidance}
However, while the standard diffusion formulation is designed to capture the comprehensive action distribution, it may generate actions that remain within this distribution but fail to meet specific task conditions, particularly in high-precision tasks such as grasping a door handle to open it.
To this end, we add additional guidance to steer its output distribution towards specific properties $y$. With the Bayes Theorem, we can introduce the given condition with an additional likelihood term $p(a_t|y)$:
\begin{equation}
     \nabla_{a_t^k} \log p(a_t^k|y) = \nabla_{a_t^k} \log p(a_t^k) + \nabla_{a_t^k} \log p(y|a_t^k).
\end{equation}

By utilizing a differentiable loss function $L(a_t^0,y)$ for sampling guidance, we eliminate the need to train the classifier or retrain the whole network. We can use the estimated likelihood of $L(a_t^0,y)$ for the additional correction step: 
\begin{equation}
\begin{aligned}
    a^{k-1}_t = \text{DDIM}(a^k_t, \boldsymbol{\epsilon}_\theta(a^k_t,k,f_t)) - \gamma \nabla_{a^k_t}L(\hat{a}{_t^0}(a_t^k), y)\,,
\end{aligned}
\end{equation}
where $\hat{a_t^0}(a_t^k)$ is an estimate based on $a_t^k$ and $\gamma$ is guidance weight.

During inference, given the robot's current joint positions $q_{pos}$, we use the forward kinematics to compute the 3D position of the robot's end-effector $p_{ee}$. We then use the distance between the end-effector's position $p_{ee}$ and the static affordance $c_{3D}$ as the adaptive loss function $\mathcal{L}_g$ for the adaptive affordance guidance. This loss function is adaptive because it is only utilized when the end-effector approaches the static affordance $c_{3D}$ within a predefined range. This prevents it from being applied when the gripper is far from or has already grasped the target object. If the gripper is approaching the target object, then the adaptive loss function can be represented as:
\begin{equation}
    \mathcal{L}_g = 
    \left\{
             \begin{array}{ll}
             \Vert p_{ee}-c_{3D} \Vert_2, &  \Vert p_{ee}-c_{3D} \Vert_2 < \theta,\\
             0, & \text{otherwise}.
             \end{array}
        \right.
\end{equation}
Here $\theta$ is a predefined threshold to control the actuating range of the affordance guidance.  Motivated by DSG~\citep{yang2023dsg}, we introduce an analytical solution to the sampling process of the diffusion model, which can guide the generation process to the direction of gradient descent while preserving the generated $a^0$ within the manifold:
\begin{equation} 
    {a_{t}^{k-1}}^* = \mu_\theta(a_{t}^{k}) - \sqrt{n}\delta_k \nabla_{a_{t}^{k}}\mathcal{L}_g /||\nabla_{a_{t}^{k}}\mathcal{L}_g||_2,
\end{equation}
\begin{equation}
\begin{aligned}
    \mu_\theta(a_{t}^{k}) = &\sqrt{\alpha_{k-1}} (a_{t}^{k}-\sqrt{1-\alpha_k}\epsilon_\theta)/\sqrt{\alpha_k} + \\
    &\sqrt{1-\alpha_{k-1}-\sigma_k^2 }\epsilon_\theta\, ,
\end{aligned}
\end{equation}
where $n$ represents the action dimensions. By incorporating affordance loss guidance into the standard diffusion formulation, AffordDP can generate action sequences directed toward the target manipulation positions for unseen objects, thereby enhancing its generalization capabilities.

\section{Experiments}

\subsection{Experiment Setup}
We evaluate our approach on multiple tasks, including both simulations and real-world scenarios. For the simulation scenario, we use IsaacGym as the simulation platform, leveraging objects from GAPartnet~\citep{geng2023gapartnet} and point clouds generated by depth cameras within the simulation for observations. 
We use the Franka arm and a side-mounted camera setting for both simulation and real-world scenarios.

\noindent\textbf{Expert demonstrations.} We use CuRoBo~\citep{sundaralingam2023curobo} as the motion planner for expert demonstration collection in the simulation. To ensure diversity in the collected trajectories, we introduce random noise to the initialization parameters, including the 6D object position, the robot arm’s base position, and the initial DOF of the robot arm. Using part bounding boxes annotated by GAPartnet, we can collect expert demonstrations specific to part manipulation. In the real-world scenario, we collect expert demonstrations through teleoperation. 

\noindent\textbf{Baseline methods.} We selected two state-of-the-art diffusion-based policies as our baselines for comparison: 1) Diffusion Policy (DP)~\cite{Chi-RSS-23}, which utilizes images as observations to generate actions via diffusion; and 2) DP3~\cite{Ze2024DP3}, which replaces image inputs in the Diffusion Policy with point clouds. We evaluate different policies by the success rate. 

\noindent\textbf{Training settings.} DP~\cite{Chi-RSS-23} and DP3~\cite{Ze2024DP3} adopt a single-object training approach, involving only one object during the training process. This setup encourages the policy to learn object-specific features, potentially limiting its generalization ability to adapt to unseen instances, which can be a common requirement in practical applications. To this end, we design two training settings: object-specific policy training and unified policy training. The former retains the original single-object training setup used in DP and DP3, while the latter employs a multi-object training approach that promotes learning task-relevant strategies for broader generalization.

\begin{figure*}
  \centering
  \includegraphics[width=0.93\linewidth]{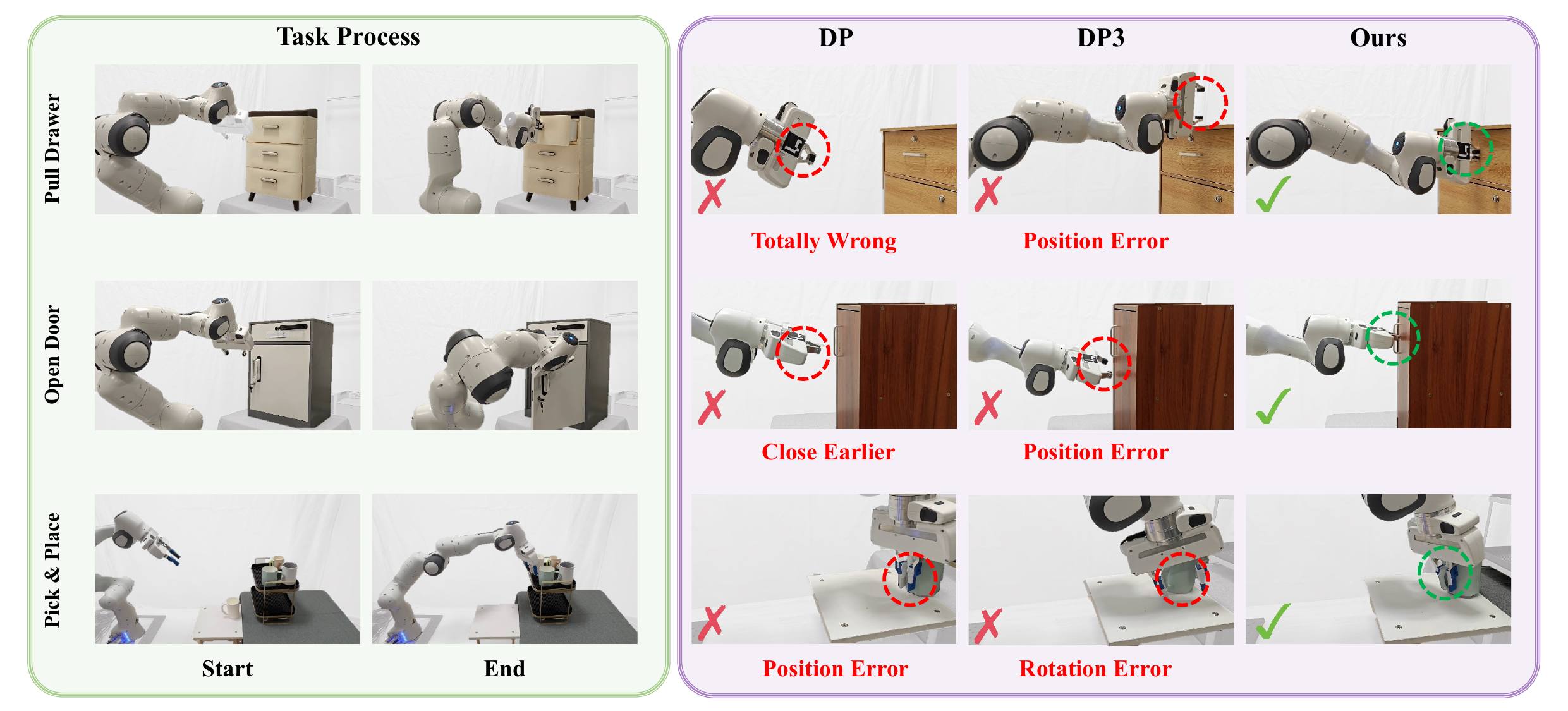}
  \caption{Policy rollouts in the real world. The left part is an intuitive demonstration of three real-world task processes, involving PullDrawer, OpenDoor, and Pick\&Place. The right part represents the policy rollout results. Without spatial perception and a lack of adequate training data, which can be over 200 demonstrations~\cite{Chi-RSS-23, Ze2024DP3}, DP fails to target the object correctly, often resulting in random and potentially unsafe actions. Lacking comprehensive static and dynamic affordance information, DP3 struggles to grasp the target accurately, even if it moves the gripper close to the target. In contrast, our method effectively grasps the target object and completes the task, demonstrating accurate spatial targeting.}
  \label{fig:task}
\end{figure*}

\subsection{Simulation Performance}
We evaluate our method on two tasks for the simulation scenario: PullDrawer and OpenDoor. For a fair comparison, we assess all methods under both training settings—object-specific policy training and unified policy training—to highlight our method's consistent advantage over baselines.

In the object-specific policy training, we collect 30 expert demonstrations for each task. To analyze how variability in training data impacts algorithm performance, we construct datasets with differing levels of variance, reflected in the initial 6D placements of the objects and the initial joint positions of the robot arm.  Our method and the baselines are evaluated across each variance level to assess robustness. The quantitative results are summarized in \cref{tab:sim-variance}. Our approach significantly outperforms the baselines across different variance levels on both tasks. 
Notably, while baseline methods are comparable to ours under low variance conditions, our method demonstrates a substantial advantage over baselines as variance levels increase.

For the unified policy training, we carefully choose 5 distinct objects for each task and collect 20 demonstrations for each object.
To evaluate generalization capabilities, we introduce three types of test objects: seen instances, unseen instances within the same category, and instances from unseen categories.
The quantitative results presented in \cref{tab:sim-result} show that our approach achieves superior performance across all three object types. Even for some categories that are not involved during training, our method still demonstrates strong generalization capabilities and exceptional performance.  In contrast, the baseline methods exhibit very low success rates with unseen categories. By leveraging transferable affordance, AffordDP can capture the inherent properties of both seen and unseen objects, significantly enhancing the policy’s learning and adaptability.

\begin{table}[h!]
\centering
\vspace{-0.05in}

\vspace{-0.05in}
\includegraphics[width=0.46\textwidth]{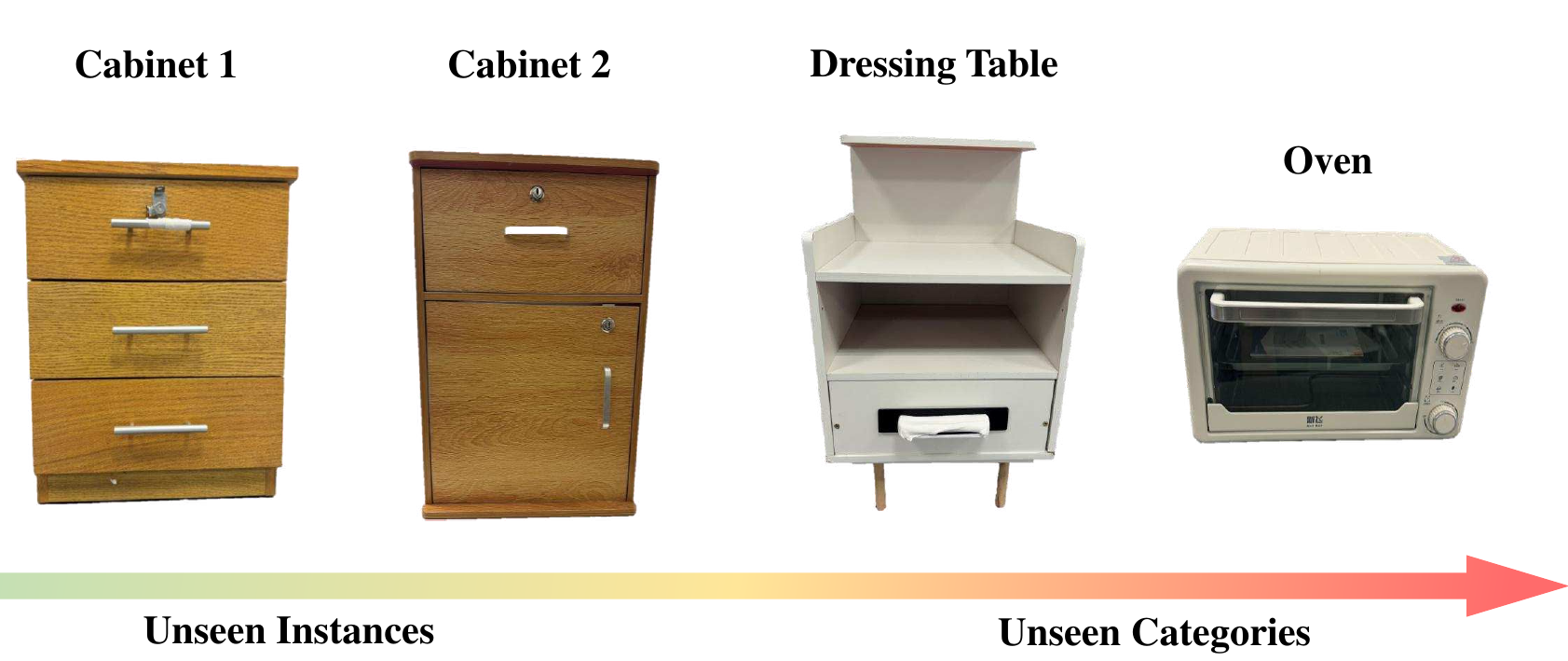}
\resizebox{0.49\textwidth}{!}{%
\begin{tabular}{l|cccccccc}
\toprule
Inst. \& Cate. Generalization & Cabinet 1 & Cabinet 2   & Dressing Table    & Oven   \\
\midrule
  DP~\cite{Chi-RSS-23} & \textcolor{gred}{\XSolidBrush} & \textcolor{gred}{\XSolidBrush} & \textcolor{gred}{\XSolidBrush} & \textcolor{gred}{\XSolidBrush}  \\
  DP3~\cite{Ze2024DP3} & \textcolor{gred}{\XSolidBrush} & \textcolor{gred}{\XSolidBrush} & \textcolor{gred}{\XSolidBrush} & \textcolor{gred}{\XSolidBrush} \\
  Ours  & \textcolor{ggreen}{\Checkmark} & \textcolor{ggreen}{\Checkmark} & \textcolor{ggreen}{\Checkmark} & \textcolor{ggreen}{\Checkmark} \\
\bottomrule
\end{tabular}}
\caption{Instance and Category Generalization on PullDrawer task. We evaluate the generalization capacities by replacing the target object with unseen instances and unseen categories. Each object undergoes three evaluation trials, and a green checkmark is awarded if the task succeeds in at least one out of the three trials.}
\label{table: instance gen}
\end{table}

\begin{table}[htb]
  \centering
  \resizebox{1.0\linewidth}{!}{
  \begin{tabular}{lcccccc}
  \hline \hline \multicolumn{7}{c}{Simulation results of object-specific policy for seen instance}  \\
    \toprule
    Tasks &\multicolumn{3}{c}{Pull Drawer} & \multicolumn{3}{c}{Open Door} \\ 
     \cmidrule(lr){2-4} \cmidrule(lr){5-7} 
    Methods & Easy & Median & Hard & Easy & Median & Hard\\

    \midrule
    DP   & \textbf{80.0\%} &  16.7\%& 13.3\% & 76.7\% & 26.7\%  & 30.0\% \\
    DP3  & \textbf{80.0\%} & 13.3\% & 16.7\% & 66.7\% & 36.7\% & 20.0\% \\
    Ours & \textbf{80.0\%} & \textbf{33.3\%}  & \textbf{26.7\%}  & \textbf{90.0\%} & \textbf{53.3\%}  & \textbf{50.0\%} \\
    \bottomrule
  \end{tabular}
  }
  \caption{Simulation performance of different methods under the object-specific policy training setting for seen instance. We categorized datasets into varying difficulty levels based on variance levels. The variance level is directly related to the random noise (position, rotation) introduced during the data collection process, the details of variance levels can be found in supplementary.}
  \label{tab:sim-variance}
  
\end{table}

\begin{table}[htb]
  \centering
  \resizebox{1.0\linewidth}{!}{
  \begin{tabular}{lcccccc}
  \hline \hline \multicolumn{7}{c}{Simulation results of the unified policy training setting} \\
    \toprule
    Tasks   &\multicolumn{3}{c}{Pull Drawer} & \multicolumn{3}{c}{Open Door} \\ 
     \cmidrule(lr){2-4} \cmidrule(lr){5-7} 
    \multirow{2}{*}{Methods} & {Seen} & Unseen& Unseen & {Seen} & Unseen & Unseen\\
     & Instance & Instance & Category & Instance & Instance & Category \\

    \midrule
    DP   & 20.7\% &  6.7\%& 3.3\% & 23.3\% & 17.5\%  & 3.3\% \\
    DP3  & 41.3\% & 10.0\% & 3.3\% & 41.1\% & 5.0\% & 5.6\% \\
    Ours & \textbf{90.0\%} & \textbf{55.6\%}  & \textbf{73.3\%}  & \textbf{74.4\%} & \textbf{52.5\%}  & \textbf{26.7\%} \\
    \bottomrule
  \end{tabular}}
  \caption{Simulation performance of different methods under the unified policy training setting. We compare our method with the baseline across three types of test objects, involving seen objects, unseen instances within the same category, and instances from unseen categories.}
  \label{tab:sim-result}
\end{table}

\begin{table*}[htb]
  \centering
  \resizebox{0.85\linewidth}{!}{
  \begin{tabular}{lcccccccccc}
  \hline \hline \multicolumn{10}{c}{ {Real-World Results} } \\
    \toprule
    {Tasks} &\multicolumn{3}{c}{{Pull Drawer}} & \multicolumn{3}{c}{{Open Door}} & \multicolumn{3}{c}{{Pick \& Place}} \\ 
     \cmidrule(lr){2-4} \cmidrule(lr){5-7} \cmidrule(lr){8-10}
    \multirow{2}{*}{{Methods}} & {Seen}& {Unseen} & {Unseen}  & {Seen} & {Unseen} & {Unseen} & Seen & {Unseen} & {Unseen} \\
     & Instance & {Instance} & {Category}  & Instance & {Instance} & {Category} & Instance & {Instance} & {Category} \\

    \midrule
    {DP} & 0.0\% & 0.0\% & 0.0\%  & 0.0\%  & 0.0\%  & 0.0\%  & 0.0\%  & 0.0\%  & 0.0\% \\
    {DP3} & 10.0\% & 0.0\% & 10.0\%  & 20.0\%  & 0.0\%  & 30.0\%  & 40.0\% & 30.0\% & 15.0\% \\
    {Ours} & \textbf{67.5\%} & \textbf{45.0\%} & \textbf{40.0\%}  & \textbf{80.0\%}  & \textbf{50.0\%}  & \textbf{50.0\%}  & \textbf{52.5\%} & \textbf{65.0\%} & \textbf{50.0\%} \\
    \bottomrule
  \end{tabular}}
  \caption{Real-World performance of different methods for all three types of test objects for three tasks. Our method consistently outperforms DP and DP3 in all scenarios, especially in tasks with unseen instances and categories. Notably, DP fails in all tasks, while DP3 achieves limited success in certain unseen categories. DP may require over 200 demonstrations for training before showing reliable performance~\cite{Chi-RSS-23, Ze2024DP3}. Observations from policy rollouts indicate that DP3 has overfitted to specific positions, allowing successful grasps only when the test object is located in those specific areas. Additionally, DP3 often produces highly overlapping action sequences across different object placements in task scenarios involving unseen instances and categories. In contrast, our method demonstrates a stable superiority. We can conclude from the task results of the seen instance that static and dynamic affordance provides a bridge across action distributions associated with different objects—something DP3 struggles to model due to the complexity of such distributions.}
  \label{real-tab}
\end{table*}

\subsection{Real-world Performance}
We evaluate our method on three tasks for the real-world scenario: OpenDoor, PullDrawer, and Pick\&Place. We use the unified policy training setting for real-world experiments. Using teleoperation, we collect 25 demonstrations per object, with a total of 4 objects for each task, resulting in 100 demonstrations per task.

Consistent with the unified policy training experiments for the simulation scenario, we evaluate our method on three types of test objects: seen instances, unseen instances within the same category, and instances from unseen categories. The results, shown in \cref{real-tab}, highlight the effectiveness of our method in real-world scenarios. 
\cref{fig:task} shows policy deployment results, showcasing our methods's superiority over baselines. And \cref{table: instance gen} gives some results on the PullDrawer task to showcase the instance and category generalization capacity of our method.
Experimental outcomes demonstrate that our method successfully generalizes to unseen instances, even those from unseen categories. In the most challenging scenarios, our method continues to exhibit strong zero-shot capabilities, significantly outperforming baseline approaches.

\vspace{-0.4em}
\begin{table}[htb]
  \centering
  \resizebox{0.95\linewidth}{!}{
  \begin{tabular}{ccc|ccc}
    \hline \hline
    Contact & \multirow{2}{*}{Trajoctary} & Affordance   & Seen & Unseen &  Unseen \\
	Point &  & Guidance  & Instance & Instance  &Category \\	
    \hline 
     \checkmark &            &            & 72.2\%  & 42.5\%  & 17.8\% \\
     \checkmark & \checkmark &            & 71.6\%  & 50.0\%  & 22.2\% \\
     \checkmark & \checkmark & \checkmark & \textbf{74.4\%}  & \textbf{52.5}\%  & \textbf{26.7\%} \\
    \bottomrule
  \end{tabular}}
  \caption{Ablation study in simulation. Success rates are shown across three scenarios: seen objects, unseen instances, and unseen categories. The combination of contact point, trajectory, and affordance guidance yields the highest success rates in both seen and unseen scenarios, particularly excelling in unseen categories.}
  \label{ablation_study}
\end{table}

\subsection{Ablation Study}
To validate the efficacy of our proposed technique, we conducted ablation studies in a simulated environment under the unified policy training setting. In these experiments, we selectively omitted two critical components, trajectory and affordance guidance, in our method while maintaining other modules unchanged, to evaluate the importance of these components.

As illustrated in \cref{ablation_study}, the performance of AffordDP markedly decreases when only the contact point is included, and the dynamic affordance is excluded. The integration of dynamic affordance leverages prior knowledge of object manipulation, thereby enhancing the model's performance, particularly for tasks with significant variability in object manipulation trajectories.

Moreover, adaptive affordance guidance substantially boosts the performance of AffordDP in previously unseen categories. 
When the model encounters objects with appearances that differ considerably from those in the training set, adaptive affordance guidance steers the generated action to gradually move toward the desired manipulation for unseen objects while keeping the generated action within the manifold of action space, effectively mitigating the impact of covariance shift.

\section{Conclusion and Discussion}
In conclusion, the Generalizable Diffusion Policy with transferable Affordance (AffordDP) significantly advances the generalization capabilities of diffusion-based robotic manipulation policies. By modeling affordances through 3D contact points and post-contact trajectories, and incorporating affordance guidance during diffusion sampling, AffordDP effectively generalizes across entirely unseen object instances and categories. The estimation of the 6D transformation matrix using foundational vision models and point cloud registration techniques allows for the transferability of these affordances. Experimental results in both simulated and real-world environments validate the superiority of AffordDP over previous methods, marking a substantial step forward in the robotic community. 

Despite these advancements, AffordDP does have some limitations that need to be addressed in future work. Current foundational models have limitations in understanding spatial information, which can restrict AffordDP’s ability to specify affordance transfers for particular parts. Additionally, AffordDP’s effectiveness is limited in scenarios where modeling or extracting affordances is challenging. Future work should focus on enhancing the spatial understanding of foundational models and improving affordance extraction techniques to overcome these limitations.

\section{Acknowledgement }
This work was supported by the Shanghai Local College Capacity Building Program (23010503100), National Natural Science Foundation of China (No.62406195), MoE Key Laboratory of Intelligent Perception and Human-Machine Collaboration  (ShanghaiTech University), HPC Platform of ShanghaiTech University and Shanghai Engineering Research Center of Intelligent Vision and Imaging. 

{
    \small
    \bibliographystyle{ieeenat_fullname}
    \bibliography{main}
}

\clearpage
\setcounter{page}{1}

\twocolumn[{
\begin{center}
{ \linespread{1.5} \selectfont
\textbf{\Large AffordDP: Generalizable Diffusion Policy with Transferable Affordance} \\
}
\Large Supplementary Materials
\end{center}
}]
\appendix

\section{Overview}

This supplementary document provides additional information, results, and visualizations to supplement the main paper. Specifically, we include:

\begin{itemize}
\item Details on data collection;
\item Information about the experimental setup;
\item Descriptions of training and inference procedures;
\item Additional experiment results.
\end{itemize}

\section{Demonstration Collection}
In this section, we provide a detailed explanation of our demonstration collection process, encompassing both simulation and real-world environments.
\subsection{Simulation Data Collection}
We employ CuRobo~\cite{sundaralingam2023curobo} as our motion planner. Given the world coordinates of the robot’s base and the target end-effector pose, CuRobo is capable of computing a feasible robotic motion trajectory. Then we utilize the bounding boxes of the annotated parts provided by GAPartnet~\cite{geng2023gapartnet} to calculate the grasping center and the pull direction post-grasping. We further plan a set of waypoints along the computed pulling direction. By combining these waypoints with the rotational components of the grasp center, we leverage CuRobo to calculate a series of feasible robot action sequences. We place an RGBD camera in the simulation environment, and the entire trajectory is recorded using this camera.

\subsection{Real-world Data Collection}
\begin{figure}[ht]
    \includegraphics[width=0.45\textwidth]{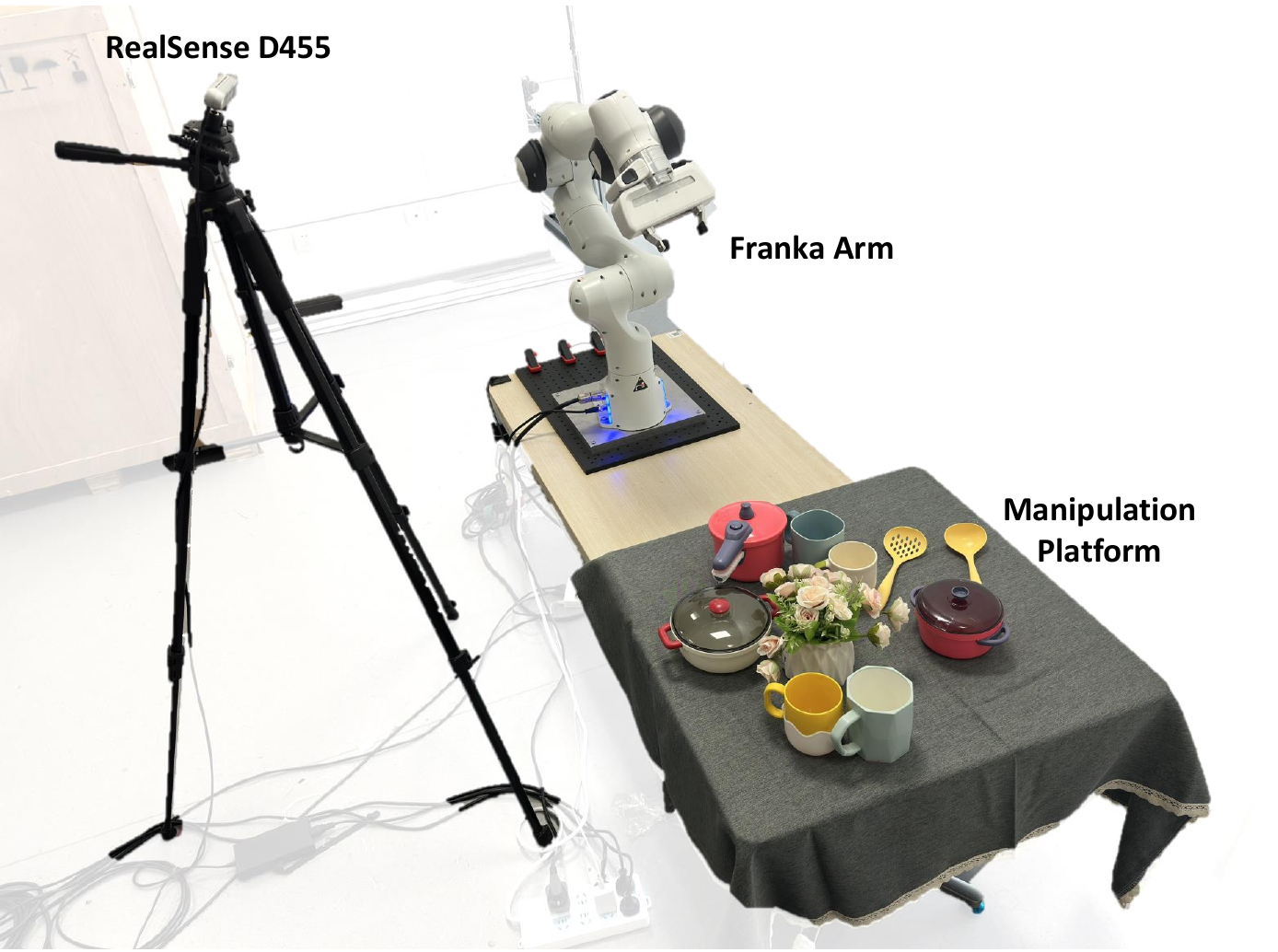}
    \caption{Real world experiment setup.}
    \label{suppl:hardware}
\end{figure}

Expert demonstrations are collected via the teleoperation system, where a human operator controls the system and the camera records the entire process, including RGB image and depth information, shown as in \cref{suppl:hardware}. We use a RealSense D455 RGBD camera to capture point cloud streams at a resolution of 640 × 480. We perform hand-eye calibration  in the real world. The calibration process enabled us to accurately determine the transformation relationship from the camera coordinate system to the robot base coordinate system. Then, we utilize the camera intrinsic parameters and this transformation matrix to convert the RGB-D images into point clouds in the robot’s base coordinate system, which facilitates subsequent policy inference.

\section{Experiment Details}
\subsection{Setup}
During the data collection process, we introduced varying levels of random noise to the initial positions and rotations of the objects, as well as to the initial position of the robotic arm. Here is the unified equation to represent the noise injection process.
\begin{equation}
    x' \mathbin{=} x + 2	\xi\mathcal{N}(0,1) - 	\xi.
\end{equation}
The magnitudes of these random noises $\xi$ correspond to different difficulty levels, as summarized in \cref{suppl:noise}.
For the unified policy training in simulation, we construct datasets using the Easy level. We summarize object categories and the number of object instances in \cref{suppl:instance}.

\begin{table}[htb]
  \centering
  \resizebox{0.75\linewidth}{!}{
  \begin{tabular}{c|ccc}
    \hline \hline
    &Hard &Median  &Easy \\ 
    \hline 
     Robot Position Noise  &0.1   &0.05   &0.025  \\
     Robot Dof Noise       &0.1   &0.05   &0.025  \\
     Object Position Noise &0.05  &0.05   &0.01  \\
    \hline
  \end{tabular}}
  \caption{Noise levels for different difficulty settings during data collection: Hard, Median, and Easy. The noise values represent the amount of random noise introduced to the robot’s position, degrees of freedom, and object position.}
  \label{suppl:noise}
\end{table}

\begin{figure*}[ht]
    \centering
    \includegraphics[width=0.95\textwidth]{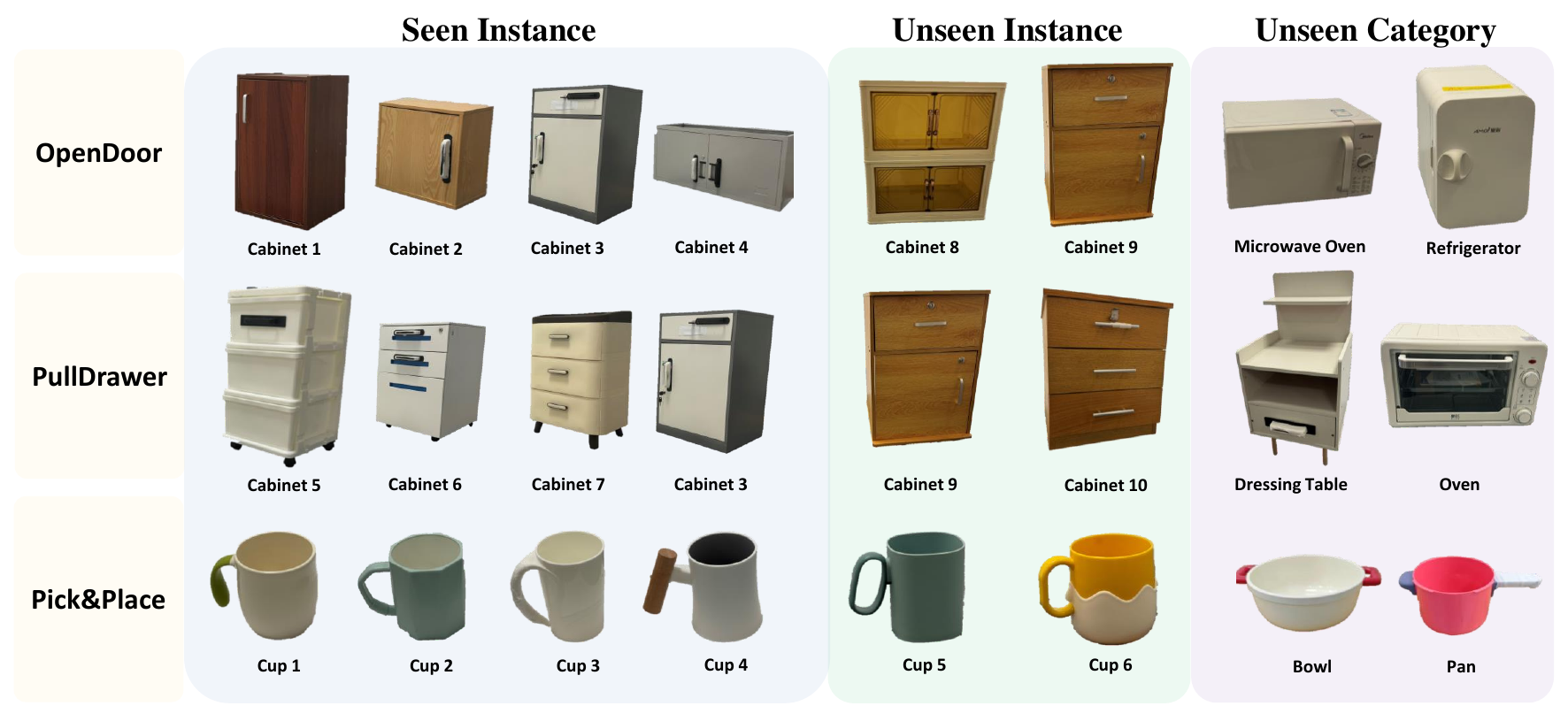}
    \caption{Objects used in Real-World Tasks, including the seen instances, unseen instances, and unseen categories.}
    \label{suppl:object}
\end{figure*}

\begin{table}[htb]
\centering
\resizebox{0.95\linewidth}{!}{
\begin{tabular}{cccc}
\hline \hline  Task Category & Task Name  & \#Train Instances & \#Test Instances \\
\hline \multirow{2}{*}{ Simulation} & OpenDoor  & 5 & 5 \\
& PullDrawer  & 5 & 5 \\
\hline \multirow{3}{*}{ Real-world } & OpenDoor & 4 & 4 \\
& PullDrawer &  4 & 4 \\ 
& Pick\&Place &  4 & 4 \\
\hline
\end{tabular}}
\caption{Number of training and testing instances for different tasks under the unified policy training setting. Simulation tasks include OpenDoor and PullDrawer, with 5 training instances and 5 testing instances each. Real-world tasks include OpenDoor, PullDrawer, and Pick\&Place, with 4 training instances and 4 testing instances each.}
\label{suppl:instance}
\end{table}
We provide additional visualizations for our objects in real-world tasks, shown in \cref{suppl:object}. We visualize all the seen instances, unseen instances, and unseen categories.

\begin{figure*}[h]
	
	\begin{minipage}{0.49\linewidth}

		\centerline{\includegraphics[width=0.85\textwidth]{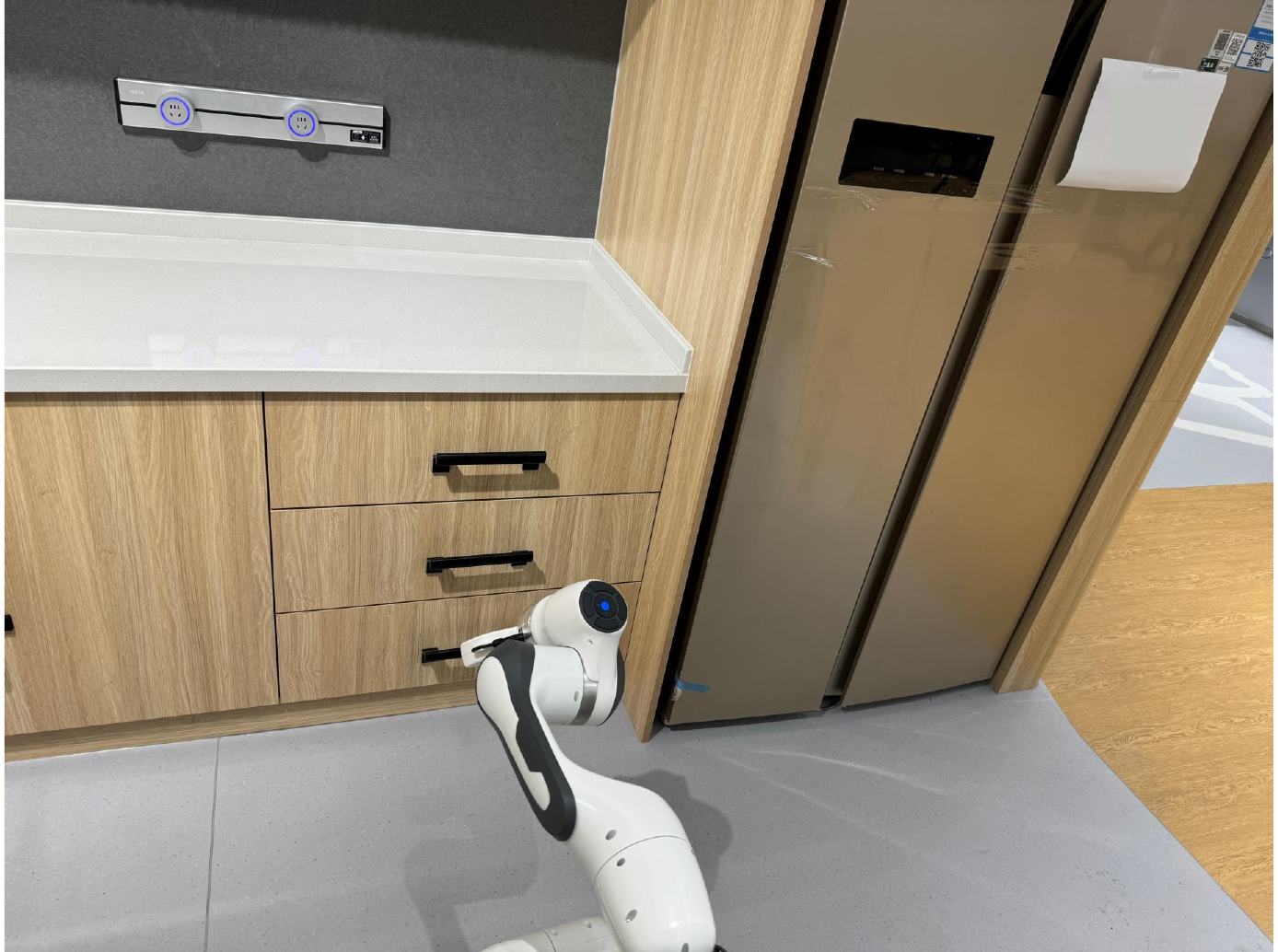}}
		\centerline{Unseen scene 1}
	\end{minipage}
	\begin{minipage}{0.49\linewidth}
		\centerline{\includegraphics[width=0.85\textwidth]{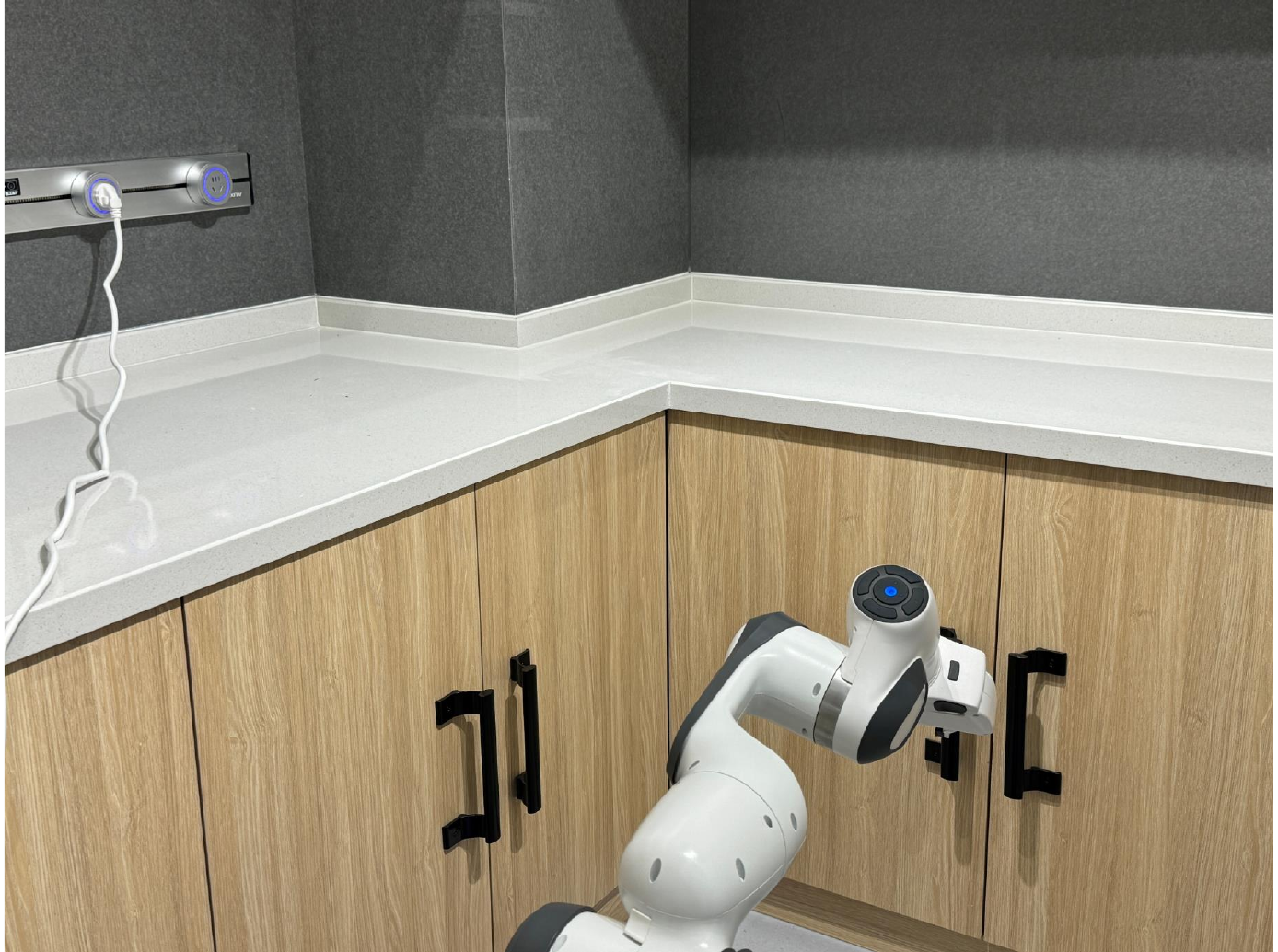}}
	 
		\centerline{Unseen scene 2}
	\end{minipage}
	\caption{Visualizations of different unseen scenes. We deployed our policy to a real-world kitchen environment in a zero-shot manner, and it still demonstrated commendable generalization. }
	\label{suppl:scene}
\end{figure*}

\subsection{Training}
We employ a convolutional network-based diffusion model as the backbone. The visual input consists of a point cloud without colors, which is downsampled from the raw point cloud using Farthest Point Sampling (FPS). We consistently use 4096 points for all simulated and real-world tasks. The representations encoded from point clouds, robot poses, and affordances are concatenated to form a unified representation with a dimension of 256. The policy in the real world is trained on a single A100 40GB GPU for two days.
Hyperparameters related to policy training are shown in \cref{suppl:Policy}.
\begin{table}[]
\centering
\resizebox{0.3\textwidth}{!}{
\begin{tabular}{lc}
\hline \hline
Hyperparameter & Default \\ \hline
Num epochs  & 4000 \\
Batch Size & 128 \\
Horizon & 16 \\
Observation Steps & 2 \\
Action Steps & 8 \\
Num points  & 4096 \\
Affordance MLP size & [64,64] \\
Affordance transformer size & 64 \\
Num attention heads & 4 \\
Num attention layers& 4 \\
Num train timesteps& 500 \\
Num inference steps & 10 \\
Learning Rate (LR) & 1.0e-4 \\
Weight decay & 1.0e-6 \\ 
\hline
\end{tabular}}
\caption{Hyperparameters of Policy Network.}
\label{suppl:Policy}
\end{table}

\subsection{Inference}
During inference, AffordDP retrieves the most similar object from the affordance memory and transfers its static and dynamic affordances to the target object. Subsequently, the affordance, visual observation, and robot proprioception are fed into the policy network to predict the action. The action is then fed into the low-level controller to operate the robot.

\begin{figure}[ht]
    \includegraphics[width=0.48\textwidth]{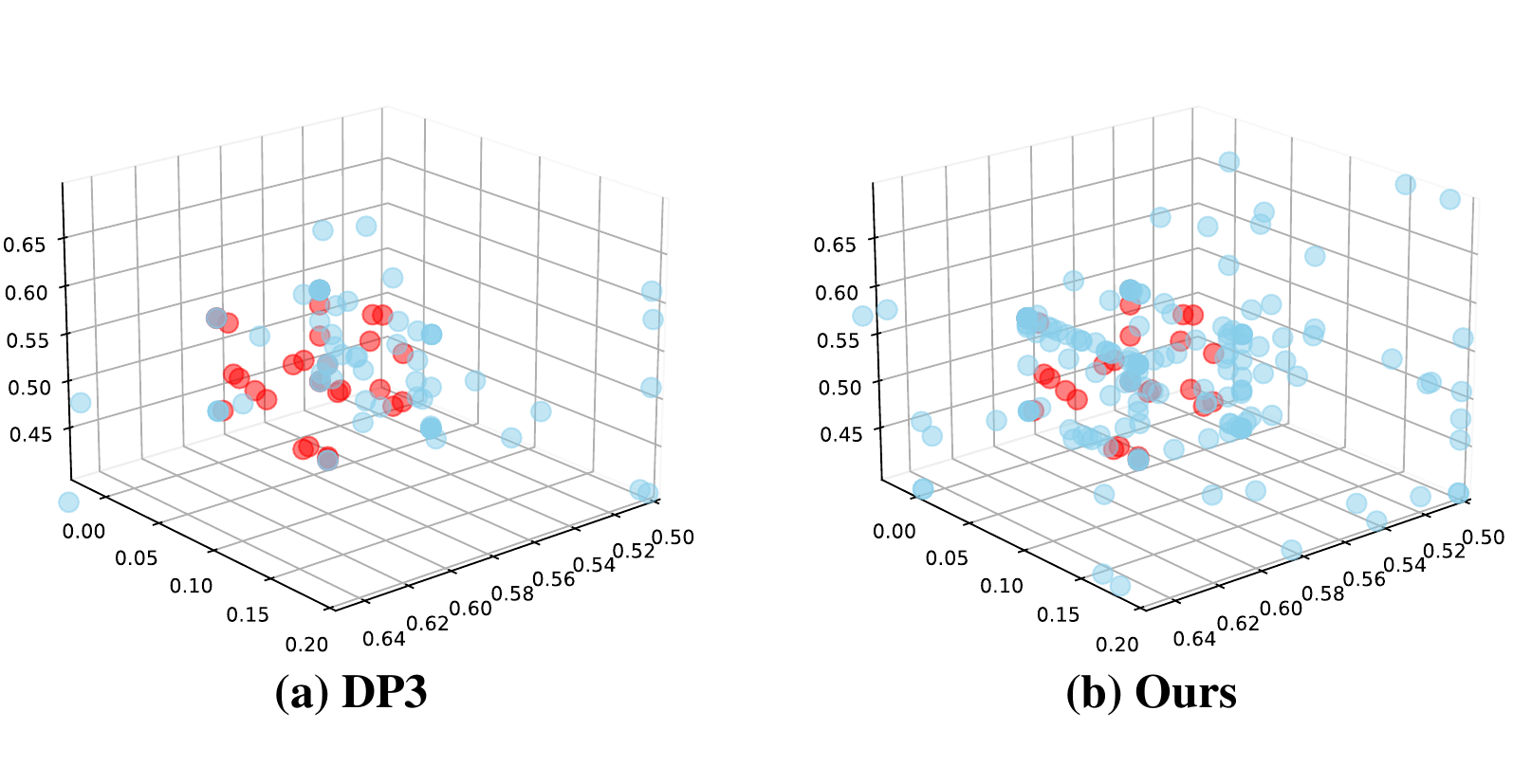}
    \caption{Spatial generalization with 30 expert demonstrations. We evaluate the performance across 1000 random positions in the 3D space. Expert demonstrations are marked as \textcolor{deepred}{$\bullet$}, and successful trials are marked as \textcolor{deepblue}{$\bullet$}. In the PullDrawer task, DP3 only succeeds in regions close to the expert demonstrations. In contrast, our method generalizes effectively, covering a broader range of 3D space, including regions not represented in the expert demonstrations.}
    \label{suppl:spatial_gen}
\end{figure}

\subsection{Evaluation}
 Here we list the evaluation criteria for each task:
 \begin{itemize}
     \item OpenDoor (sim): The task is considered successful if the robotic arm accurately grasps the specific door handle and opens the door to an angle of 30 degrees.
     \item PullDrawer (sim): The task is considered successful if the robotic arm accurately grasps the specific drawer handle and extends the drawer by a distance of 0.15 meters.
     \item OpenDoor: The task is considered successful if the robotic arm accurately grasps a specific door handle and actuates the door to a predetermined angular displacement.
     \item PullDrawer: The task is considered successful if the robotic arm accurately grasps a specific drawer handle and extends the drawer along a defined linear distance.
     \item Pick\&Place: The task is considered successful if the robotic arm accurately grasps a specific region of the object and places the object onto a designated storage rack.
 \end{itemize}

\section{Additional experiment results}
\subsection{Spatial Generalization}
Following DP3~\cite{Ze2024DP3}, we utilized PullDrawer as our benchmark task in simulation, providing 30 demonstrations (visualized by \textcolor{deepred}{$\bullet$}). We randomly initialized the positions of the objects and conducted 1,000 evaluations (successful positions are visualized by \textcolor{deepblue}{$\bullet$}), shown in \cref{suppl:spatial_gen}. The number of successful trials using DP3 is 65, whereas our approach achieved 170 successful trials.

\subsection{More Real World Experiments}
We give the real-world results for each object in \cref{tab:suppl:results}, which is supplementary to \cref{real-tab} in our main paper.

\begin{table*}[htbp]
\begin{flushleft}
\resizebox{0.95\textwidth}{!}{
\begin{tabular}{l|cccc|cc|cc}
\hline \hline
& \multicolumn{7}{c}{Open Door} \\

 Methods& Cabinet 1  & Cabinet 2  & Cabinet 3 & Cabinet 4 &  Cabinet 8 & Cabinet 9 & Microwave Oven & Refrigerator\\
\midrule

DP   & 0/10 & 0/10 & 0/10  & 0/10 & 0/10 & 0/10 & 0/10 & 0/10\\
DP3  & 3/10 & 2/10 & 3/10  & 0/10 & 0/10 & 0/10 & \textbf{6/10} & 0/10\\
Ours & \textbf{6/10} & \textbf{8/10} & \textbf{10/10} & \textbf{8/10} & \textbf{6/10} & \textbf{4/10} & \textbf{6/10} & \textbf{4/10}\\
\hline
\end{tabular}}
\vspace{5pt}

\resizebox{0.9\textwidth}{!}{
\begin{tabular}{l|cccc|cc|cc}
\hline \hline
& \multicolumn{7}{c}{Pull Drawer} \\

 Methods& Cabinet 5  & Cabinet 6  & Cabinet 7 & Cabinet 3 & Cabinet 9 & Cabinet 10 & Dressing Table  & Oven\\
\midrule

DP   & 0/10 & 0/10 & 0/10 & 0/10 & 0/10 & 0/10 & 0/10 & 0/10 \\
DP3  & 1/10 & 0/10 & 0/10 & 3/10 & 0/10 & 0/10 & 2/10 & 0/10 \\
Ours & \textbf{6/10} & \textbf{5/10} & \textbf{7/10} & \textbf{9/10} & \textbf{5/10} & \textbf{4/10} & \textbf{5/10} & \textbf{3/10} \\
\hline
\end{tabular}}

\vspace{5pt}

\begin{tabular}{l|cccc|cc|cc}
\hline \hline
& \multicolumn{7}{c}{Pick\&Place} \\

 Methods& \ \ Cup 1\ \   &\ \  Cup 2\ \   & \ \ Cup 3\ \ &\ \ Cup 4\ \ &\ \ Cup 5\ \ &\ \ Cup 6\ \ &\ \ Bowl\ \ &\ \ Pan\ \ \\
\midrule
DP   & 0/10 & 0/10 & 0/10 & 0/10 & 0/10 & 0/10 & 0/10 & 0/10 \\
DP3  &\textbf{6/10} & 3/10 & 2/10 & \textbf{5/10} & 3/10 & 3/10 & 0/10 & 3/10\\
Ours & \textbf{6/10} & \textbf{6/10} & \textbf{5/10} & 4/10 & \textbf{6/10} & \textbf{7/10} & \textbf{4/10} & \textbf{6/10}\\
\hline
\end{tabular}
\caption{Success rates of different methods on real-world tasks. We evaluated the performance of different methods on various objects.}
\label{tab:suppl:results}
\end{flushleft}
\end{table*}

\subsection{Semantic correspondence model selection}
\begin{figure*}[ht]
    \centering
    \includegraphics[width=0.95\textwidth]{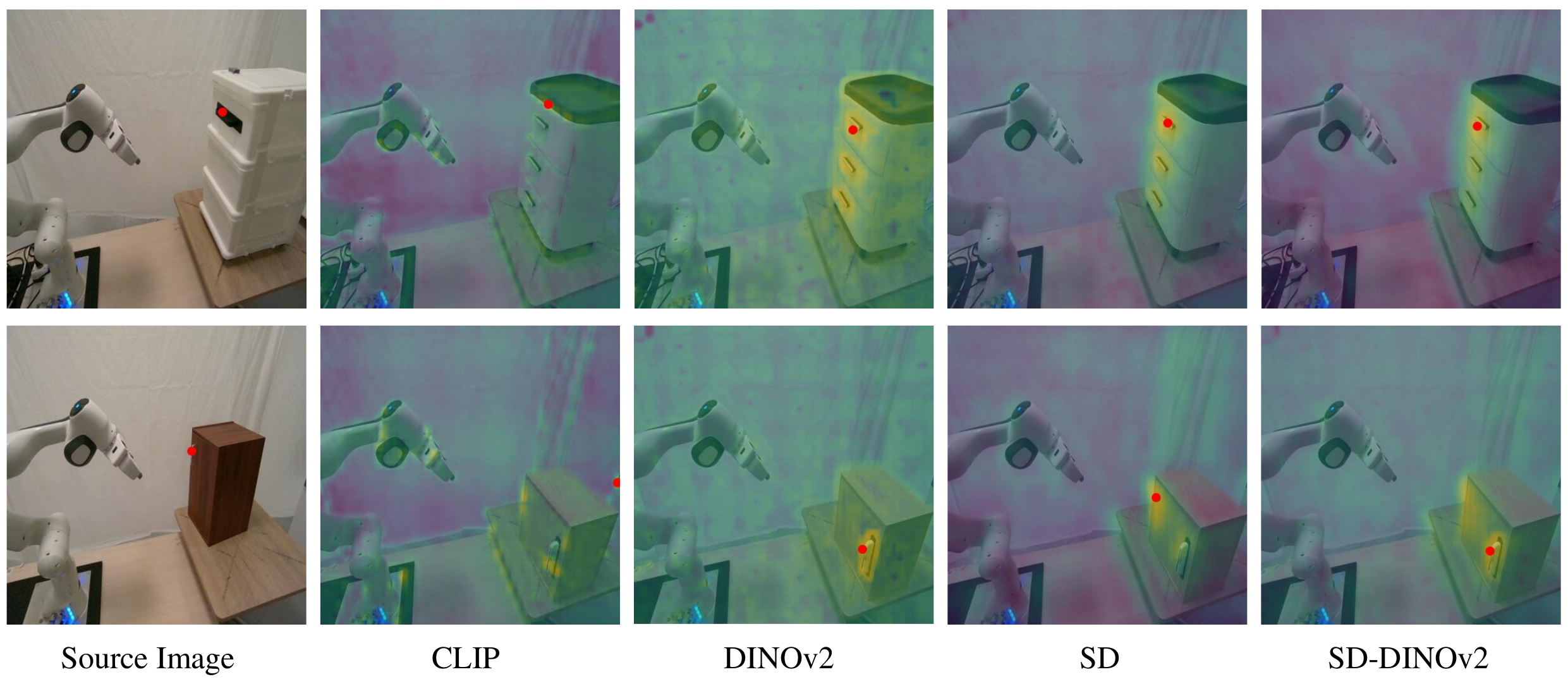}
    \caption{Comparison of semantic correspondence transfer among different foundational models. Red points \textcolor{red}{$\bullet$} represent the static affordance and its corresponding transferred results.}
    \label{suppl:semantic}
\end{figure*}

We compared several foundational vision models (CLIP, DINOv2, SD, and SD-DINOv2) and provided qualitative results on semantic correspondence transfer, shown as \cref{suppl:semantic}. The lack of sufficient semantic understanding in CLIP can lead to incorrect semantic correspondence transfer. Similarly, the SD model also exhibits transfer errors in certain tasks. In comparison to DINOv2 and SD, the SD-DINOv2 model demonstrates greater stability and exhibits smaller errors in the transfer of static affordance within the pixel space.

\subsection{Unseen scene generalization}
To further demonstrate the generalization capability of our method, we applied the policy zero-shot transfer to unseen scenes(kitchen environment), as shown in \cref{suppl:scene}. We recalibrated the camera and cropped the extraneous points from the point cloud to execute our policy. Surprisingly, despite the stark contrast with our training scene, our method still demonstrated robust generalization capabilities. Please refer to our project website for detailed videos.

\section{Limitations}
The limitations of our method are mainly two-fold: scenarios where foundation models fail, such as severe occlusion or visual distortion; and tasks requiring precise force control where affordance extraction is difficult, like grasping eggs without breaking it.

\end{document}